\documentclass[preprint,12pt]{elsarticle}

\usepackage{amsmath,amssymb,mathtools}
\usepackage{booktabs}
\usepackage{graphicx}
\usepackage{float}
\usepackage{microtype}
\usepackage{array}
\usepackage{multirow}
\usepackage{tabularx}
\usepackage{url}
\usepackage{xcolor}
\usepackage{enumitem}
\usepackage{hyperref}
\hypersetup{hidelinks,pdftitle={The Stochastic Gap: A Markovian Framework for Pre-Deployment Reliability and Oversight-Cost Auditing in Agentic Artificial Intelligence},pdfauthor={Biplab Pal; Santanu Bhattacharya}}
\newcommand{\E}{\mathbb{E}}
\newcommand{\Pp}{\mathbb{P}}
\newcommand{\1}{\mathbf{1}}

\newcommand{\mcC}{\mathcal{C}}
\newcommand{\mcS}{\mathcal{S}}
\newcommand{\mcA}{\mathcal{A}}
\newcolumntype{L}[1]{>{\raggedright\arraybackslash}p{#1}}

\begin{document}

\begin{frontmatter}

\title{The Stochastic Gap: A Markovian Framework for Pre-Deployment Reliability and Oversight-Cost Auditing in Agentic Artificial Intelligence}

\author[umbc]{Biplab Pal}
\ead{bpal1@umbc.edu}

\author[mit]{Santanu Bhattacharya\corref{cor1}}
\ead{santanu2@mit.edu}

\cortext[cor1]{Corresponding author.}
\address[umbc]{CARDS (Center for Real-Time Distributed Sensing and Autonomy), University of Maryland Baltimore County, Baltimore, MD, USA}
\address[mit]{Massachusetts Institute of Technology, Cambridge, MA, USA}

\begin{abstract}
Agentic artificial intelligence (AI) in organizations is a sequential decision problem constrained by reliability and oversight cost. When deterministic workflows are replaced by stochastic policies over actions and tool calls, the key question is not whether a next step appears plausible, but whether the resulting trajectory remains statistically supported, locally unambiguous, and economically governable. We develop a measure-theoretic Markov framework for this setting. The core quantities are state blind-spot mass $B_n(\tau)$, state-action blind mass $B^{SA}_{\pi,n}(\tau)$, an entropy-based human-in-the-loop escalation gate, and an expected oversight-cost identity over the workflow visitation measure.

We instantiate the framework on the Business Process Intelligence Challenge 2019 purchase-to-pay log (251{,}734 cases, 1{,}595{,}923 events, 42 distinct workflow actions) and construct a log-driven simulated agent from a chronological 80/20 split of the same process. The main empirical finding is that a large workflow can appear well supported at the state level while retaining substantial blind mass over next-step decisions: refining the operational state to include case context, economic magnitude, and actor class expands the state space from 42 to 668 and raises state-action blind mass from 0.0165 at $\tau=50$ to 0.1253 at $\tau=1000$. On the held-out split, $m(s)=\max_a \hat\pi(a\mid s)$ tracks realized autonomous step accuracy within 3.4 percentage points on average.

The same quantities that delimit statistically credible autonomy also determine expected oversight burden. The framework is demonstrated on a large-scale enterprise procurement workflow and is designed for direct application to engineering processes for which operational event logs are available.
\end{abstract}

\begin{keyword}
Agentic artificial intelligence \sep Pre-deployment auditing \sep Markov decision processes \sep Human-in-the-loop oversight \sep Oversight cost \sep Artificial intelligence safety \sep Off-policy evaluation \sep Autonomous workflow reliability
\end{keyword}

\end{frontmatter}

\section{Introduction}\label{sec:intro}

Public discussion of agentic artificial intelligence (AI) has been dominated by prompting, orchestration, and demonstrations of short-horizon competence. Enterprise deployment, however, confronts a different problem. Operational workflows are engineered to behave near-deterministically through approval rules, validation checks, separation of duties, and exception-handling logic. Once a large language model (LLM)-driven or agentic policy is inserted into that control structure, execution is no longer described by one-step plausibility alone; it is described by a trajectory distribution over a constrained process. The central question is therefore whether a policy can sustain reliable \emph{paths} through a workflow, not merely whether it can produce a locally plausible next action.

This mismatch is no longer hypothetical. Air Canada's customer-service chatbot supplied incorrect bereavement-fare guidance, and the firm was subsequently found liable for negligent misrepresentation \cite{brand2025aircanada}. McDonald's ended its International Business Machines (IBM) voice-ordering pilot after mixed operational results and persistent ordering errors \cite{granthamphilips2024mcdonalds}. At the portfolio level, Gartner projected in 2025 that more than 40\% of agentic-AI projects would be canceled by 2027 because of escalating costs, unclear business value, or inadequate risk controls \cite{gartner2025agentic}; IBM reported that only 25\% of AI initiatives had delivered expected return on investment and only 16\% had scaled enterprise-wide \cite{ibm2025roi}. These observations suggest that reliability and cost are not secondary deployment details. They are coequal design constraints.

We refer to the resulting analytical problem as the \emph{stochastic gap}. In a deterministic enterprise resource planning (ERP) or rule-based system, the next state is effectively a function of the current state. In an agentic system, the next action is sampled or selected from a policy and the workflow evolves under a transition kernel. Local uncertainty, rare branches, and process loops can then accumulate along the path. The practical question is not whether an agent can act, but how much autonomy the historical process can actually justify.

The contribution of this paper is a Markovian framework that makes that question auditable from event-log data. The framework has four parts. First, we define state blind-spot mass and state-action blind mass as finite-sample analogues of deployment mass that lies outside or near the edge of historical support. Second, we add Shannon entropy and a reproducible risk weighting to obtain a deployment-side escalation rule for human intervention. Third, we show that the same gate that induces a reliability envelope also induces an expected oversight-cost identity, thereby coupling reliability and economics at the level of workflow visitation. Fourth, we validate the theory on a genuine multi-step enterprise workflow. We use the full Business Process Intelligence Challenge 2019 (BPI 2019) purchase-to-pay log for the descriptive audit and then build a chronological held-out \emph{simulated agent} from the same process to compare theoretical reliability surrogates against realized step and case outcomes.

The resulting picture is more specific than the slogan that ``agents need human-in-the-loop (HITL) oversight.'' It shows where historical support is adequate, where branching ambiguity forces escalation, how blind mass reappears once next-step decisions are modeled explicitly, and how the induced escalation policy determines the cost of deployment. In this sense, the framework addresses the two concerns that now dominate enterprise agent adoption: whether the system can be trusted, and whether required human oversight leaves deployment economically viable.

\begin{table}[H]
\centering
\small
\caption{Motivating industry evidence for a joint reliability-and-cost framing of agentic artificial intelligence.}
\label{tab:industry}
\begin{tabularx}{\textwidth}{@{}L{0.21\textwidth}L{0.47\textwidth}L{0.24\textwidth}@{}}
\toprule
Context & Observed issue & Why it matters here \\
\midrule
Air Canada chatbot \cite{brand2025aircanada} & Customer-facing chatbot provided incorrect fare guidance; liability attached to the enterprise rather than to the model artifact. & Reliability failures propagate into legal and governance risk. \\
McDonald's--IBM drive-through pilot \cite{granthamphilips2024mcdonalds} & Large-scale voice-ordering pilot ended after mixed results and repeated ordering errors. & Sequential operational reliability, not one-shot fluency, governs real deployment. \\
Gartner 2025 agentic-AI forecast \cite{gartner2025agentic} & Over 40\% of agentic-AI projects projected to be canceled because of escalating costs, unclear value, or inadequate risk controls. & Reliability and cost must be analyzed jointly rather than separately. \\
IBM 2025 ROI guidance \cite{ibm2025roi} & Only 25\% of AI initiatives reported as delivering expected ROI; only 16\% scaled enterprise-wide. & Human oversight and integration cost can dominate the business case. \\
\bottomrule
\end{tabularx}
\end{table}

\section{Related work and positioning}\label{sec:related}

Recent work on language agents has focused on architectures that interleave reasoning and action, use external tools, or coordinate multiple agent roles. ReAct couples reasoning traces with actions in interactive environments \cite{yao2023react}; Toolformer learns application programming interface (API) invocation from self-supervision \cite{schick2023toolformer}; AutoGen provides a multi-agent conversation framework \cite{wu2023autogen}; Reflexion and CRITIC add self-reflection and tool-assisted critique to improve multi-step behavior \cite{shinn2023reflexion,gou2023critic}. This literature establishes that multi-step agentic behavior is technically feasible, but it does not determine \emph{where} enterprise autonomy is statistically justified before deployment.

That gap is visible in evaluation work as well. AgentBench, SWE-bench, OSWorld, and $\tau$-bench show that current agents remain brittle on long-horizon reasoning, software engineering, open computer environments, and tool use under policy constraints \cite{liu2023agentbench,jimenez2024swebench,xie2024osworld,yao2024taubench}. Recent safety work further emphasizes indirect prompt injection, tool misuse, and the difficulty of constraining autonomous behavior in open settings \cite{greshake2023prompt,amodei2016concrete,bai2022constitutional}. These benchmarks and safety studies expose failure modes after an agent has been defined. Our question is earlier and more deployment-facing: given the event log already available inside an organization, which parts of the workflow are supported enough to justify autonomy at all?

The mathematical background comes from constrained sequential decision making. Safe reinforcement learning and constrained policy optimization formalize how performance objectives interact with explicit constraints \cite{garcia2015safe,achiam2017cpo}. Logged-bandit and off-policy-evaluation work show that value estimation is impossible without adequate support overlap between the target and logging policies \cite{swaminathan2015crm,dudik2011dr,jiang2016dr}. Selective prediction and conformal prediction show how abstention can be made principled when the model is uncertain \cite{geifman2019selectivenet,angelopoulos2023conformal}. Our framework adapts that logic to enterprise workflows by using support, entropy, and risk to delimit an \emph{autonomy envelope} prior to deployment.

Process-mining research supplies the event-log formalism that makes such audits operational in organizations. Event logs record case-wise trajectories, resources, timestamps, and control-flow variants, and they are routinely used to reconstruct and evaluate business processes \cite{vanderaalst2016}. Conformance checking detects deviations between observed traces and normative process models \cite{carmona2018conformance}; predictive process monitoring anticipates future violations or outcomes from partial traces \cite{teinemaa2019outcome}. Our contribution is complementary. Rather than measuring conformance after specifying a normative process or forecasting case outcomes from partial traces, we ask where a candidate autonomous policy is adequately supported, sufficiently low-entropy, and inexpensive enough in oversight burden to justify routine execution.

Finally, the economics literature on generative AI documents that productivity gains are real but heterogeneous. Improvements have been reported in writing-intensive work and customer-support settings, while field evidence also points to a jagged technological frontier in which AI materially helps some tasks and degrades others \cite{noy2023experimental,brynjolfsson2025genai,dellacqua2023jagged}. Industry surveys suggest that ROI realization remains limited at scale \cite{ibm2025ceo,ibm2025roi}. The present paper contributes a missing layer in that discussion: a workflow-level audit that ties reliability and oversight cost to the same empirical quantities.

\section{Enterprise workflows as Markov systems}\label{sec:markov}

Let a case $c \in \mcC$ generate a trajectory
\begin{equation}
\gamma^c = (s_0^c,a_0^c,s_1^c,a_1^c,\ldots,s_{T_c}^c),
\end{equation}
where $s_t^c \in \mcS$ is a workflow state and $a_t^c \in \mcA$ is the chosen next workflow action. Under a fixed abstraction, the process can be treated as a finite-state Markov decision process. In general, the workflow state $s_t$ is a tuple that captures the current activity, case-level context variables (such as document status or verification flags), a discretized measure of economic magnitude, and the class of actor responsible for the decision. The specific variables used to instantiate each dimension are domain-dependent and should be chosen to reflect the control-relevant context of the deployed process.

For the BPI 2019 purchase-to-pay log, these dimensions correspond to
\begin{equation}
\label{eq:state}
 s_t = \bigl(e_t,\; \mathrm{item\_type}_c,\; \mathrm{GR}_c,\; \mathrm{value\_bin}_t,\; \mathrm{actor\_class}_t\bigr),
\end{equation}
where $e_t$ is the current activity, $\mathrm{GR}_c$ is the goods-receipt-based invoice-verification flag, $\mathrm{value\_bin}_t$ is a discretized bin of absolute cumulative net worth, and $\mathrm{actor\_class}_t\in\{\text{human},\text{system}\}$ is derived from the log resource field. Actions correspond to next recorded activities such as \texttt{Record Goods Receipt}, \texttt{Record Invoice Receipt}, or \texttt{Remove Payment Block}. Section~\ref{sec:data} describes how alternative domains would parameterize the same state template differently.

Given counts $N(s,a)$ in a training or logging sample, the empirical next-step policy is
\begin{equation}
\hat{\pi}(a\mid s)=\frac{N(s,a)}{N(s)}, \qquad N(s)=\sum_{a}N(s,a),
\end{equation}
and the empirical transition kernel is
\begin{equation}
\hat{P}(s'\mid s,a)=\frac{N(s,a,s')}{N(s,a)}.
\end{equation}
The reliability object is a path probability. If $\Gamma$ denotes trajectory space and $\mathrm{Safe}\subseteq \Gamma$ denotes admissible trajectories, then
\begin{equation}
R(\pi)=\Pp_{\pi}\{\Gamma\in\mathrm{Safe}\}.
\end{equation}
This makes the core deployment question explicit: path reliability depends on which parts of trajectory space are statistically supported and which are not.

\begin{figure}[t]
\centering
\includegraphics[width=0.88\textwidth]{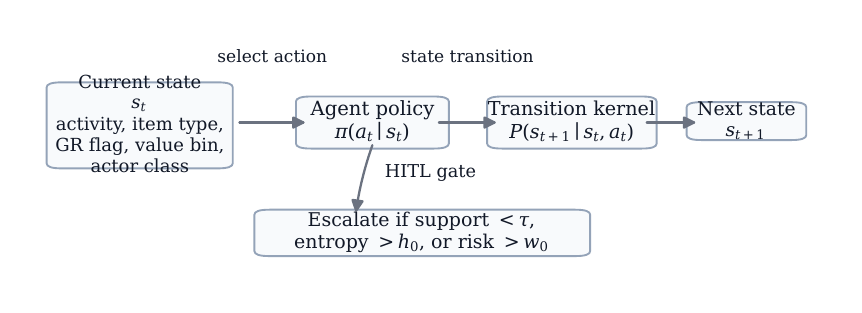}
\caption{Markov reliability model for scoped autonomy in enterprise workflows. A workflow state is mapped to an action through policy $\pi(a_t\mid s_t)$, the process evolves under transition kernel $P(s_{t+1}\mid s_t,a_t)$, and a human-in-the-loop (HITL) gate escalates states with inadequate support, high branching entropy, or elevated risk.}
\label{fig:model}
\end{figure}

\begin{figure}[t]
\centering
\includegraphics[width=0.93\textwidth]{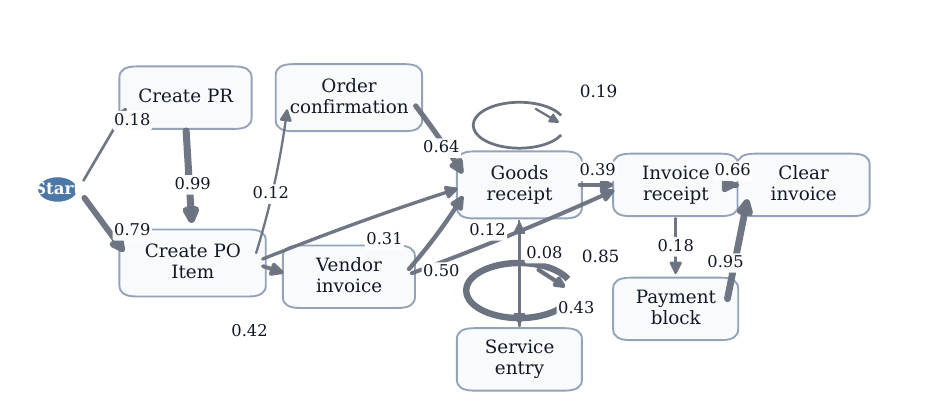}
\caption{Dominant empirical transitions in the BPI 2019 purchase-to-pay process. Edge widths are proportional to empirical transition probabilities. The log contains recurrent loops and exception-handling branches; the mean case length is 6.34 events, the 99th percentile is 24 events, the maximum observed case length is 990, and the transition-level self-loop rate is 15.7\%.}
\label{fig:transitions}
\end{figure}

Figure~\ref{fig:model} shows the conceptual control model used throughout the paper. Figure~\ref{fig:transitions} shows the dominant transition subgraph extracted from the full BPI 2019 log. The data are genuinely sequential: 251{,}734 cases generate 1{,}595{,}923 events, the mean case length is 6.34 events, the median is 5, the 99th percentile is 24, and the longest observed case contains 990 events. Moreover, 15.7\% of all transitions are self-loops and 6.3\% of cases contain at least one self-loop, so path inflation and rework are not edge cases.

\section{Reliability and oversight-cost framework}\label{sec:framework}

\subsection{Blind-spot mass over states and actions}

Let $d_{\pi}(s)$ denote the deployment occupancy measure over states and $d_{\pi}(s,a)$ the deployment occupancy measure over state-action pairs. In practice we estimate these from the empirical log:
\begin{equation}
\hat d_{\pi}(s)=\frac{N_n(s)}{n}, \qquad \hat d_{\pi}(s,a)=\frac{N_n(s,a)}{\sum_{u,b}N_n(u,b)}.
\end{equation}
For a minimum-support threshold $\tau \ge 1$, define the state blind-spot mass
\begin{equation}
B_n(\tau)=\sum_{s}\hat d_{\pi}(s)\,\1\{N_n(s)<\tau\},
\end{equation}
and the state-action blind mass
\begin{equation}
B^{SA}_{\pi,n}(\tau)=\sum_{s,a}\hat d_{\pi}(s,a)\,\1\{N_n(s,a)<\tau\}.
\end{equation}
The first quantity measures how much deployment mass lies in low-support states. The second is more directly relevant to agentic systems because an agent must choose a next action rather than merely recognize a state.

If $P$ denotes the deployment measure and $Q$ the logging or training measure, the measure-theoretic issue is support mismatch. When $P \not\ll Q$, the Lebesgue decomposition
\begin{equation}
P = P_{\mathrm{ac}} + P_{\perp}
\end{equation}
splits deployment mass into a component absolutely continuous with the log and a singular component that cannot be recovered from it. The finite-sample quantities $B_n(\tau)$ and $B^{SA}_{\pi,n}(\tau)$ are operational analogues of this mismatch: even when $P \ll Q$, materially important parts of the deployed process can remain under-supported.

For any step-level correctness indicator $C_t \in \{0,1\}$ and supported set
\begin{equation}
S_{\tau}=\{(s,a):N_n(s,a)\ge \tau\}, \qquad U_{\tau}=S_{\tau}^{c},
\end{equation}
we have the decomposition
\begin{equation}
\label{eq:coverage_ceiling}
\E[C_t] = \bigl(1-B^{SA}_{\pi,n}(\tau)\bigr)\E[C_t\mid (s,a)\in S_{\tau}] + B^{SA}_{\pi,n}(\tau)\E[C_t\mid (s,a)\in U_{\tau}].
\end{equation}
Equation~\eqref{eq:coverage_ceiling} is the sequential version of the coverage ceiling: strong performance on supported transitions cannot by itself justify strong performance on blind transitions.

\subsection{Entropy, risk weighting, and human-in-the-loop (HITL) gating}

Support is not the only issue. Even in well-supported states, the next-step distribution may be diffuse. We quantify local ambiguity with Shannon entropy,
\begin{equation}
H\bigl(\hat\pi(\cdot\mid s)\bigr) = -\sum_{a}\hat\pi(a\mid s)\log_2 \hat\pi(a\mid s).
\end{equation}

Let $V_t$ denote the cumulative net worth recorded for event $t$. For the autonomy gate we define a state-level value-risk score
\begin{equation}
\label{eq:w_state}
\bar V(s)=\frac{1}{N(s)}\sum_{t:s_t=s}|V_t|,
\qquad
w(s)=\frac{\log\!\bigl(1+\bar V(s)\bigr)}{\max_{u}\log\!\bigl(1+\bar V(u)\bigr)} \in [0,1].
\end{equation}
To summarize decision-level consequence we also define a pairwise risk score
\begin{equation}
\label{eq:w_pair}
w_{SA}(s,a)=\frac{1}{N(s,a)}\sum_{t:(s_t,a_t)=(s,a)}\left[0.6\,\frac{\log(1+|V_t|)}{\max_{u}\log(1+|V_u|)} + 0.4\,\1\{a_t\in E_{\mathrm{exc}}\}\right],
\end{equation}
where $E_{\mathrm{exc}}$ is the exception-sensitive activity set listed in Appendix~A. The corresponding risk-weighted transition blind mass is
\begin{equation}
\label{eq:blind_risk}
B^{SA,*}_{\pi,n}(\tau)=\sum_{s,a}\hat d_{\pi}(s,a)\,w_{SA}(s,a)\,\1\{N_n(s,a)<\tau\}.
\end{equation}
A simple escalation gate is then
\begin{equation}
\label{eq:gate}
G_{\tau,h_0,w_0}(s)=\1\{N(s)<\tau\; \lor\; H(\hat\pi(\cdot\mid s))>h_0\; \lor\; w(s)>w_0\}.
\end{equation}
We summarize the gate with the event-level and case-level autonomous shares,
\begin{align}
A_{\mathrm{event}} &= 1 - \sum_s \hat d_\pi(s) G_{\tau,h_0,w_0}(s),\\
A_{\mathrm{case}} &= \frac{1}{|\mcC|}\sum_{c\in\mcC}\1\{G_{\tau,h_0,w_0}(s_t^c)=0,\;\forall t<T_c\}.
\end{align}
$A_{\mathrm{event}}$ measures the share of nonterminal decisions that remain autonomous. $A_{\mathrm{case}}$ measures the share of complete workflows that remain autonomous from start to finish. Their difference quantifies path compounding.

\subsection{Oversight cost and case-level reliability surrogates}

The same gate that delimits reliability also determines operational cost. Let
\begin{equation}
D_{\pi}(s)=\E_{\pi}\Bigl[\sum_{t=0}^{T-1}\1\{s_t=s\}\Bigr]
\end{equation}
be the expected visit-count measure under policy $\pi$, where $T$ denotes the number of nonterminal decisions in a case. If an autonomous decision incurs unit cost $c_A$ and an escalated decision incurs unit cost $c_H>c_A$, then the expected operational cost per case under gate $G$ is
\begin{equation}
\label{eq:cost_identity}
\mathcal{C}(\pi;G)=c_A\E[T] + (c_H-c_A)\sum_{s} D_{\pi}(s)G(s).
\end{equation}
Thus the same statistics that delimit the autonomy envelope also determine expected oversight burden. Adding an error penalty $\lambda>0$ yields
\begin{equation}
\label{eq:cost_error}
\mathcal{C}_{\lambda}(\pi;G)=\mathcal{C}(\pi;G)+\lambda\E_{\pi}\Bigl[\sum_{t=0}^{T-1}(1-G(s_t))(1-C_t)\Bigr].
\end{equation}
Equations~\eqref{eq:cost_identity}--\eqref{eq:cost_error} provide a direct bridge between reliability and economics: a more permissive autonomy envelope lowers escalation cost but increases exposure to autonomous error.

For a deterministic greedy agent $\hat a(s)=\arg\max_a \hat\pi(a\mid s)$, define
\begin{equation}
\label{eq:mstate}
m(s)=\max_a \hat\pi(a\mid s).
\end{equation}
Under the usual conditional-independence surrogate along a held-out path, the probability of zero-touch completion is approximated by
\begin{equation}
\label{eq:zero_touch_surrogate}
\widetilde C_0(G)=\E\Bigl[\prod_{t=0}^{T-1}(1-G(s_t))\,m(s_t)\Bigr],
\end{equation}
and the corresponding safe-completion surrogate with human-in-the-loop fallback is
\begin{equation}
\label{eq:safe_surrogate}
\widetilde R_{\mathrm{safe}}(G)=\E\Bigl[\prod_{t=0}^{T-1}\bigl(G(s_t)+(1-G(s_t))m(s_t)\bigr)\Bigr].
\end{equation}
These quantities are conservative whenever repeated visits to the same gateway states or structured backbone routes create positive dependence among correct actions. We use them below as theoretical surrogates and compare them against realized outcomes of a held-out log-driven agent.

\section{Data, agent construction, and evaluation protocol}\label{sec:data}

Although the empirical instantiation below uses a procurement workflow, the state template in Section~\ref{sec:markov} is intended to be domain-generic: other deployments would populate the same slots with domain-specific control variables, economic context, and actor classes.

The BPI 2019 event log records purchase-to-pay traces from a multinational company and is publicly available through 4TU.ResearchData \cite{bpi2019data,bpi2019challenge}. Following standard event-log ordering practice \cite{vanderaalst2016}, events were ordered by case identifier, timestamp, and event identifier. The resulting full log contains 251{,}734 cases, 1{,}595{,}923 events, 42 distinct workflow actions, and both human and system users.

The process starts most often at \texttt{Create Purchase Order Item} (79.4\% of cases) or \texttt{Create Purchase Requisition Item} (18.5\%). Dominant transitions include Create Purchase Requisition Item $\rightarrow$ Create Purchase Order Item (0.99), Order Confirmation $\rightarrow$ Record Goods Receipt (0.64), and Record Invoice Receipt $\rightarrow$ Clear Invoice (0.66). At the same time, the log contains strong recurrent structure: \texttt{Record Service Entry Sheet} self-loops with probability 0.85 and \texttt{Record Goods Receipt} self-loops with probability 0.19.

To measure state refinement in the descriptive audit we compare three abstractions:
\begin{align}
 s_t^{(1)} &= (e_t),\\
 s_t^{(2)} &= (e_t,\mathrm{item\_type}_c,\mathrm{GR}_c),\\
 s_t^{(3)} &= (e_t,\mathrm{item\_type}_c,\mathrm{GR}_c,\mathrm{value\_bin}_t,\mathrm{actor\_class}_t).
\end{align}
Value bins are derived from the empirical distribution of absolute cumulative net worth and correspond to zero, low, mid, high, and very-high transaction intensity. The refined state therefore captures process-control context, economic context, and the human-versus-system distinction.

For the agent study we use a chronological 80/20 split by case completion time. The training segment contains 201{,}387 cases and 1{,}267{,}250 events; the held-out segment contains 50{,}347 cases and 328{,}673 events. The simulated agent is intentionally simple: on a held-out state $s$, it escalates if $G_{\tau,h_0,w_0}(s)=1$ and otherwise chooses the greedy action $\hat a(s)=\arg\max_a \hat\pi(a\mid s)$. Because counterfactual next states are unavailable in observational logs, correctness is evaluated with respect to the observed next activity in the held-out trace. This yields an imitation-style but genuinely sequential evaluation.

Let $a_t^{\mathrm{obs}}$ denote the next activity observed in the held-out case. The realized zero-touch completion of the held-out agent is
\begin{equation}
\label{eq:actual_zero}
C_0^{\mathrm{test}}(G)=\frac{1}{|\mcC_{\mathrm{test}}|}\sum_{c\in\mcC_{\mathrm{test}}}\prod_{t=0}^{T_c-1}\1\{G(s_t^c)=0\}\1\{\hat a(s_t^c)=a_t^{\mathrm{obs},c}\},
\end{equation}
and the realized safe completion with human fallback is
\begin{equation}
\label{eq:actual_safe}
R_{\mathrm{safe}}^{\mathrm{test}}(G)=\frac{1}{|\mcC_{\mathrm{test}}|}\sum_{c\in\mcC_{\mathrm{test}}}\prod_{t=0}^{T_c-1}\Bigl[\1\{G(s_t^c)=1\}+\1\{G(s_t^c)=0\}\1\{\hat a(s_t^c)=a_t^{\mathrm{obs},c}\}\Bigr].
\end{equation}
We also report the mean number of human touches per case,
\begin{equation}
\label{eq:touches}
H_{\mathrm{case}}^{\mathrm{test}}(G)=\frac{1}{|\mcC_{\mathrm{test}}|}\sum_{c\in\mcC_{\mathrm{test}}}\sum_{t=0}^{T_c-1}G(s_t^c),
\end{equation}
which is the empirical analogue of the escalation term in Eq.~\eqref{eq:cost_identity}.

\section{Results}\label{sec:results}

\subsection{Full-log audit: state coverage is not the whole story}

Table~\ref{tab:abstractions} summarizes the descriptive audit on the full BPI 2019 log. Refining the operational state from activity only to activity+item type+GR flag+value bin+actor increases the observed state count from 42 to 668 and the observed state-action count from 498 to 3262. Coarse state occupancy remains comparatively well covered at this scale: even under the refined abstraction, $\hat B_n(200)=0.0108$ and $\hat B_n(1000)=0.0460$.

The support limitation becomes visible at the sequential layer. For the refined abstraction, $\hat B^{SA}_{\pi,n}(50)=0.0165$, $\hat B^{SA}_{\pi,n}(200)=0.0462$, and $\hat B^{SA}_{\pi,n}(1000)=0.1253$. Thus more than 12\% of observed transition mass remains in state-action pairs with fewer than one thousand historical examples even though the full log contains 1.6 million events. The risk-weighted blind mass follows the same pattern, with $\hat B^{SA,*}_{\pi,n}(200)=0.0202$ and $\hat B^{SA,*}_{\pi,n}(1000)=0.0505$. The implication is that support over \emph{actions}, rather than over states alone, is the quantity that governs where agentic execution is statistically justified.

\begin{table}[t]
\centering
\footnotesize
\caption{State refinement and blind mass on the full BPI 2019 log. Coarse state occupancy can appear well covered even when materially under-supported regions persist at the state-action layer.}
\label{tab:abstractions}
\resizebox{\textwidth}{!}{%
\begin{tabular}{@{}lrrrrrr@{}}
\toprule
Abstraction & States & SA pairs & $B_n(200)$ & $B_n(1000)$ & $B^{SA}_{\pi,n}(200)$ & $B^{SA}_{\pi,n}(1000)$ \\
\midrule
Activity only & 42 & 498 & 0.0004 & 0.0021 & 0.0077 & 0.0324 \\
Activity + item type + GR & 190 & 1217 & 0.0020 & 0.0212 & 0.0194 & 0.0681 \\
Activity + item type + GR + value + actor & 668 & 3262 & 0.0108 & 0.0460 & 0.0462 & 0.1253 \\
\bottomrule
\end{tabular}%
}
\end{table}

\begin{figure}[t]
\centering
\includegraphics[width=0.94\textwidth]{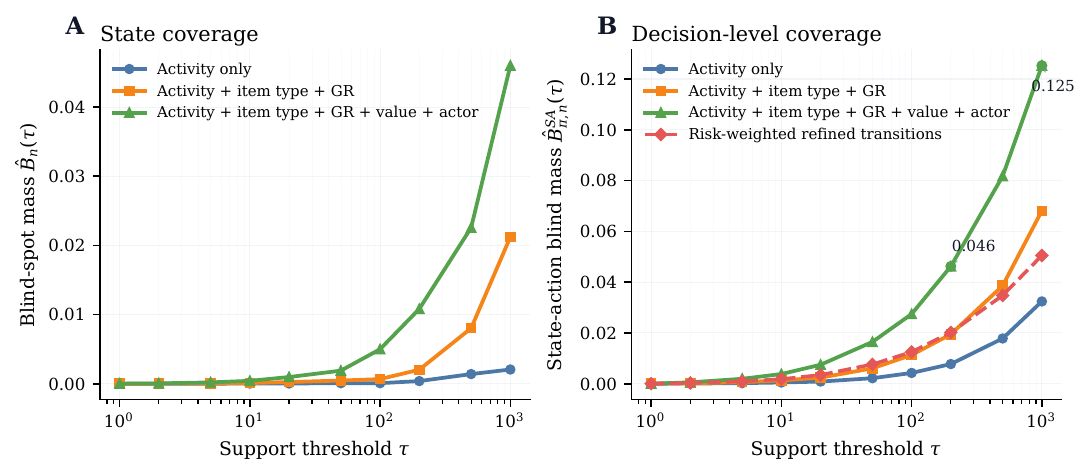}
\caption{Coverage on the full BPI 2019 log. (A) State blind-spot mass $\hat B_n(\tau)$ remains modest even under refined state abstractions. (B) State-action blind mass $\hat B^{SA}_{\pi,n}(\tau)$ increases substantially faster, particularly once value and actor context are included. The dashed red curve reports the corresponding risk-weighted blind mass for the refined transition model.}
\label{fig:coverage}
\end{figure}

\subsection{Full-log audit: entropy concentrates the autonomy bottleneck}

Coverage is not sufficient on its own. Under the refined full-log abstraction, the highest-entropy states are concentrated in human-handled approval and exception-management contexts: approval changes for standard items without GR-based verification in the high-value bin ($H=3.07$ bits, $N=1379$), delivery-indicator changes for standard items without GR-based verification in the low-value bin ($H=3.01$ bits, $N=573$), and price changes for standard items without GR-based verification in the very-high-value bin ($H=2.88$ bits, $N=658$). At the activity level, the largest next-step entropies occur for approval changes (3.19 bits), goods-receipt cancellations (2.96 bits), and delivery-indicator changes (2.88 bits), whereas the service-entry region is comparatively predictable once contextual variables are included.

\begin{figure}[t]
\centering
\includegraphics[width=0.70\textwidth]{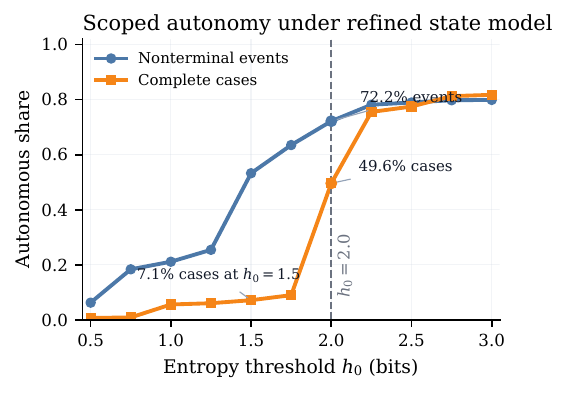}
\caption{Scoped autonomy on the full BPI 2019 log under the refined state abstraction with $N(s)\ge 50$ and $w(s)\le 0.6$. Event-level autonomy remains substantially higher than end-to-end case-level autonomy because local ambiguity compounds along the trajectory.}
\label{fig:autonomy}
\end{figure}

Figure~\ref{fig:autonomy} converts these local entropies into an operational autonomy envelope for the full log. With $\tau=50$ and $w_0=0.6$, increasing the entropy allowance $h_0$ expands the fraction of steps eligible for autonomous execution. At $h_0=2.0$ bits, 72.2\% of nonterminal events satisfy the gate, but only 49.6\% of complete cases remain fully autonomous. At the stricter threshold $h_0=1.5$, the event-level autonomous share remains 53.3\%, whereas the end-to-end case-level share drops to 7.1\%. The path effect is therefore substantial: a workflow can look mostly autonomous at the step level while remaining only partially autonomous from start to finish.

\subsection{Held-out agent validation: theoretical surrogates track realized behavior}

The descriptive audit above is retrospective. We now ask whether the same quantities are predictive of the realized behavior of a held-out agent. The simulated agent is trained on the first 80\% of cases in chronological order and evaluated on the final 20\% using the refined state representation. Table~\ref{tab:agent} reports representative thresholds.

At the step level, the theoretical surrogate $m(s)=\max_a \hat\pi(a\mid s)$ tracks realized autonomous accuracy closely. Across the full range $h_0 \in [1.0,3.0]$, the mean absolute gap between theoretical and realized autonomous step accuracy is 3.4 percentage points, with a maximum gap of 4.0 percentage points. At $h_0=2.0$, for example, the surrogate predicts 67.5\% autonomous step accuracy and the held-out agent achieves 63.7\%.

At the case level, the surrogate $\widetilde R_{\mathrm{safe}}$ is conservative but directionally accurate. With $\tau=50$ and $w_0=0.6$, the held-out agent achieves 55.6\% safe case completion at $h_0=1.5$, 49.6\% at $h_0=2.0$, and 45.2\% at $h_0=2.25$, while mean human touches per case fall from 3.02 to 2.26 and then to 1.90. The surrogate underestimates these values because Eq.~\eqref{eq:safe_surrogate} treats correct autonomous actions as conditionally independent along the path. In the actual process, once a case remains on the low-entropy backbone, later decisions become positively correlated and success exceeds the product approximation.

The same mechanism explains the striking jump in complete-case autonomy around $h_0=2.0$. In the temporal split, 33 gateway states satisfy the support and risk conditions while lying in the entropy band $1.5 < H(\hat\pi(\cdot\mid s)) \le 2.0$. These states account for 13.4\% of held-out nonterminal decisions but are visited by 69.1\% of held-out cases. Most are procurement hubs centered on \texttt{Create Purchase Order Item}, \texttt{Record Goods Receipt}, and \texttt{Record Invoice Receipt}. Once these hubs are admitted into the autonomous envelope, a large fraction of cases no longer encounter any escalation trigger, and case-level autonomy rises sharply.

Finally, zero-touch completion remains substantially below case autonomy. At $h_0=2.0$, 42.3\% of held-out cases stay entirely inside the autonomous envelope, but only 16.1\% are completed with no escalation and no autonomous mismatch. This gap quantifies how much reliability is lost when autonomy is widened without corresponding improvement in local predictive certainty.

\begin{table}[t]
\centering
\small
\caption{Held-out agent validation on the chronological 80/20 split of BPI 2019 under the refined state abstraction, with $\tau=50$ and $w_0=0.6$. $\widetilde R_{\mathrm{safe}}$ is the theoretical surrogate from Eq.~\eqref{eq:safe_surrogate}; $R_{\mathrm{safe}}^{\mathrm{test}}$ is the realized safe completion with human fallback from Eq.~\eqref{eq:actual_safe}; $C_0^{\mathrm{test}}$ is realized zero-touch completion from Eq.~\eqref{eq:actual_zero}.}
\label{tab:agent}
\begin{tabular}{@{}rrrrrrr@{}}
\toprule
$h_0$ & $\widetilde m_{\mathrm{step}}$ & $m_{\mathrm{step}}^{\mathrm{test}}$ & $\widetilde R_{\mathrm{safe}}$ & $R_{\mathrm{safe}}^{\mathrm{test}}$ & $C_0^{\mathrm{test}}$ & touches/case \\
\midrule
1.50 & 0.721 & 0.684 & 0.520 & 0.556 & 0.033 & 3.02 \\
2.00 & 0.675 & 0.637 & 0.404 & 0.496 & 0.161 & 2.26 \\
2.25 & 0.654 & 0.615 & 0.354 & 0.452 & 0.231 & 1.90 \\
\bottomrule
\end{tabular}
\end{table}

\begin{figure}[t]
\centering
\includegraphics[width=0.96\textwidth]{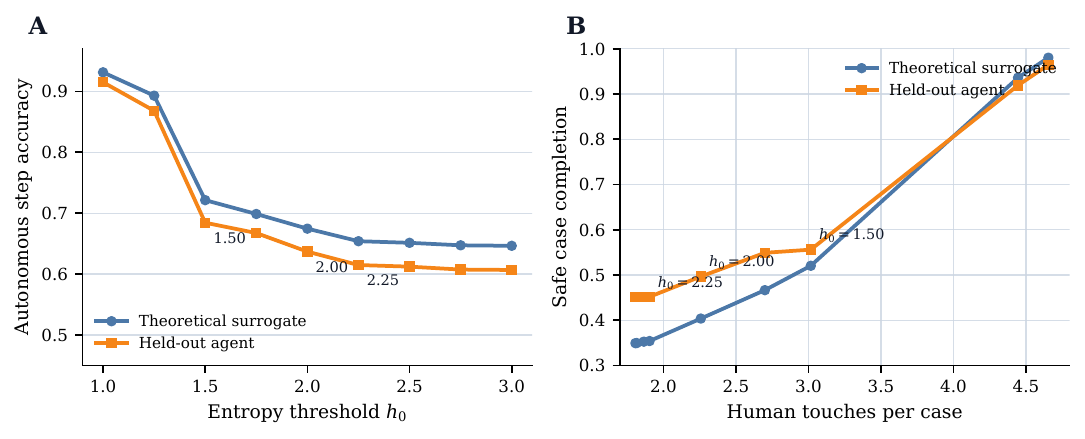}
\caption{Theory-versus-agent comparison on the chronological held-out split of BPI 2019. (A) Theoretical autonomous-step surrogate $m(s)=\max_a \hat\pi(a\mid s)$ versus realized held-out step accuracy as the entropy threshold $h_0$ varies, with mean absolute gap 3.4 percentage points. (B) Reliability-cost frontier induced by the same gate: the x-axis is mean human touches per case, the y-axis is safe case completion, and the theoretical curve is conservative but monotone relative to the held-out agent.}
\label{fig:validation}
\end{figure}

\subsection{Reliability and cost are coupled, not separable}

The empirical comparison shows that reliability and cost cannot be optimized independently. Stricter gates increase safe completion but at the price of heavier oversight. More permissive gates lower oversight burden but transfer probability mass to autonomous decisions made in higher-entropy regions. This is precisely the behavior predicted by Eqs.~\eqref{eq:cost_identity}--\eqref{eq:cost_error}. The important point is that the same state statistics drive both terms: support deficiency and branching ambiguity increase the probability of escalation, while also lowering the realized accuracy of any agent that is allowed to act without review.

In this sense, the framework is useful even when the organization has no deployed agent yet. Before prompting, fine-tuning, or tool integration, the event log already determines an empirical reliability-cost frontier. The frontier does not answer every causal question, but it constrains the plausible design space: it identifies where full autonomy is unjustified, where human-in-the-loop oversight is economically dominant, and where additional training data or workflow redesign would be most valuable.

\section{Discussion and limitations}\label{sec:discussion}

A first limitation is that the BPI log is observational rather than agent-generated. This prevents direct counterfactual evaluation of arbitrary actions and means that the held-out agent study is imitation-style: correctness is assessed relative to the next action realized in the historical trace. That is appropriate for the pre-deployment question studied here, but it is not a substitute for online evaluation of a deployed system.

A second limitation concerns state representation. The analysis uses a first-order Markov approximation after state abstraction. Under the maintained assumption that added variables are semantically aligned with the deployed control state, enriching the representation with longer memory, latent tool state, retrieved documents, or dialogue history will usually expand the effective state space and make support harder rather than easier. In that calibrated sense, the reported blind masses are lower bounds on the support problem faced by richer agent implementations. A misspecified refinement that fragments the state along irrelevant dimensions could, however, reduce or distort blind mass numerically, so the lower-bound interpretation is conditional rather than automatic.

A third limitation concerns the risk proxy. The weight $w(s)$ is designed to be reproducible from publicly available fields in the BPI log and therefore emphasizes value intensity and exception-sensitive activities. In a real deployment, the risk functional should be domain-specific and should incorporate policy criticality, compliance burden, and downstream error cost more directly.

Finally, the industry evidence in Table~\ref{tab:industry} is motivational rather than exhaustive. The point is not that any single case settles the scientific question, but that practical deployments already encounter the two pressures captured by the framework: reliability failures create governance exposure, while oversight requirements and integration burden create cost pressure. The scientific contribution of the paper is to make those pressures measurable from event-log data before deployment.

\section{Conclusion}\label{sec:conclusion}

Agentic artificial intelligence in enterprise settings should not be analyzed as a prompting problem alone. It is a stochastic-control problem in which reliability and oversight cost are jointly determined by support, branching ambiguity, and path structure. The relevant object is not an isolated next step, but the state-action trajectory through a constrained workflow.

Using the full BPI 2019 purchase-to-pay log, this paper shows that a large enterprise process can appear well covered at the state level while still retaining nontrivial blind mass over next-step decisions. Using a chronological held-out split of the same log, it further shows that the resulting theoretical quantities are predictive of realized agent behavior: state-wise top-probability is a good surrogate for held-out step accuracy, and the gate-induced reliability-cost frontier remains conservative but directionally accurate at the case level.

Operationally, the result supports a simple recommendation. Enterprise agent deployment should begin with a support, entropy, risk, and oversight-cost audit rather than with prompt iteration alone. Within that framework, the stochastic gap becomes measurable, and so does the amount of autonomy that can be justified.

\appendix
\section{Exception-sensitive activity set used in the BPI risk weights}\label{sec:risk_appendix}

For the risk-weighted transition audit in Eq.~\eqref{eq:blind_risk}, the exception-sensitive activity set is
\begin{quote}\small
Change Approval for Purchase Order; Change Delivery Indicator; Change Final Invoice Indicator; Change Price; Change Quantity; Change Rejection Indicator; Change Storage Location; Change payment term; Delete Purchase Order Item; Remove Payment Block; Set Payment Block; Cancel Goods Receipt; Cancel Invoice Receipt; Cancel Subsequent Invoice; Vendor creates debit memo; Block Purchase Order Item; Reactivate Purchase Order Item; Update Order Confirmation.
\end{quote}
The code bundle accompanying the manuscript computes Eqs.~\eqref{eq:w_state}--\eqref{eq:blind_risk} directly from the public BPI 2019 log using this definition.

\section*{Data and code availability}
The event log analyzed in this paper is publicly available as the BPI Challenge 2019 dataset through 4TU.ResearchData \cite{bpi2019data}. The package accompanying this manuscript contains the full \LaTeX{} source, all figure assets, and Python scripts that rebuild the descriptive and agent-validation figures from the public event log.

\section*{Funding}
This research did not receive any specific grant from funding agencies in the public, commercial, or not-for-profit sectors.

\section*{Declaration of competing interest}
The authors declare that they have no known competing financial interests or personal relationships that could have appeared to influence the work reported in this paper.

\section*{Declaration of generative AI and AI-assisted technologies in the writing process}
During the preparation of this work, the authors used a large language model to assist with language editing, organization of the manuscript, and code scaffolding for figure generation. The authors reviewed, edited, and verified the content and take full responsibility for the final manuscript.

\end{document}